\documentclass{article}

\usepackage{microtype}
\usepackage{graphicx}
\usepackage{subcaption}
\usepackage{booktabs} 

\usepackage{times}
\usepackage[T1]{fontenc}

\usepackage{latexsym}
\usepackage{graphicx}
\usepackage{amsfonts}
\usepackage{multirow}
\usepackage{arydshln}
\usepackage{amsmath}
\usepackage{amssymb}
\usepackage{float}
\usepackage{pgf}
\usepackage{pgfplots}
\usepackage{tikz}
\usepackage{tikzscale}

\usepackage{microtype}

\include{plots}
\definecolor{bblue}{HTML}{4F81BD}
\definecolor{rred}{HTML}{C0504D}
\definecolor{ggreen}{HTML}{9BBB59}
\definecolor{ppurple}{HTML}{9F4C7C}
\definecolor{Dark scarlet}{HTML}{560319}
\definecolor{Forest green}{HTML}{1E4D2B}

\usepackage{hyperref}

\usepackage[accepted]{icml2020}

\icmltitlerunning{Explainable and Discourse Topic-aware Neural Language Understanding}

\begin{document}
	
	\twocolumn[
	\icmltitle{Explainable and Discourse Topic-aware Neural Language Understanding}
	
	
	
	\icmlsetsymbol{equal}{*}
	
	\begin{icmlauthorlist}
		\icmlauthor{Yatin Chaudhary}{siemens,lmu}
		\icmlauthor{Hinrich Sch\"utze}{lmu}
		\icmlauthor{Pankaj Gupta}{siemens}
	\end{icmlauthorlist}
	
	\icmlaffiliation{siemens}{Corporate Technology, Machine Intelligence (MIC-DE), Siemens AG, Munich, Germany}
	\icmlaffiliation{lmu}{CIS, University of Munich (LMU), Munich, Germany}
	
	\icmlcorrespondingauthor{Yatin Chaudhary}{yatin.chaudhary@drimco.net}
	
	\icmlkeywords{Machine Learning, ICML}
	
	\vskip 0.3in
	]
	
	
	
	\printAffiliationsAndNotice{}  
	
	\begin{abstract}
		Marrying topic models and language models exposes language understanding to a broader source of document-level context beyond sentences via topics.  
		While introducing topical semantics in language models, existing approaches incorporate latent document topic proportions and ignore topical discourse in sentences of the document. 
		This work extends the line of research by additionally introducing an explainable topic representation in language understanding, obtained from a set of key terms correspondingly for each latent topic of the proportion.   
		Moreover, we retain sentence-topic association along with document-topic association by modeling topical discourse for every sentence in the document.  
		We present a novel neural composite language modeling (NCLM) framework that exploits both the latent and explainable topics along with topical discourse at sentence-level in a joint learning framework of topic and language models. 
		Experiments over a range of tasks such as language modeling, word sense disambiguation, document classification, retrieval and text generation demonstrate ability of the proposed model in improving language understanding.
	\end{abstract}
	
	\section{Introduction} \label{sec:introduction}
	
	Topic models (TMs) such as 
	LDA ~\citep{DBLP:conf/nips/BleiNJ01}
	facilitate document-level semantic knowledge in the form of topics, explaining the thematic structures hidden in a document collection. In doing so, they learn document-topic association in a generative fashion by counting word-occurrences across documents. Essentially, the generative framework assumes that each document is a mixture of latent topics, i.e., topic-proportions and each latent topic is a unique distribution over words in vocabulary.   
	Beyond a document representation, topic models also offer interpretability via topics (a set of top key terms).  Recently, neural topic models \cite{DBLP:conf/iclr/GuptaCBS19, DBLP:conf/aaai/GuptaCBS19, DBLP:conf/icml/MiaoYB16} have been shown to outperform LDA-based models. Thus, we consider neural network based topic models in this work. 
	
	Language models (LMs) \cite{DBLP:conf/interspeech/MikolovKBCK10, DBLP:conf/naacl/PetersNIGCLZ18} have recently gained success in natural language understanding by predicting the next (target) word in a sequence given its preceding and/or following context(s), accounting for linguistic structures such as word ordering. However, LM are often contextualized by an n-gram window or a sentence, ignoring  global semantics in context beyond the sentence boundary especially in modeling documents. To capture long-term semantic dependencies, recent works \cite{DBLP:conf/aistats/WangGWSHPSC18, DBLP:conf/acl/LauBC17, DBLP:conf/iclr/Dieng0GP17} have attempted to introduce document-level semantics in LMs at sentence-level by marrying topic and language models, e.g., augmenting LSTM-based LMs with a latent document-topic proportion (association) obtained from a topic model for the document in which the sentence appears.   
	
	\begin{figure*}[t]
		\begin{center}
			\includegraphics[scale=0.96]{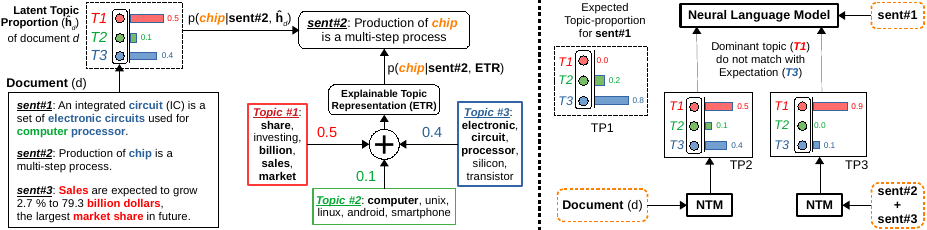}
			\caption{Detailed illustration of: (a)  Motivation \#1: top 5 key terms of each topic provide a fine-grained outlook of document semantics context than $\mathbf{h}_d$ for prediction of word ``chip''; (b) Motivation \#2: Negative influence via sentence-level topical discourse mismatch.}
			\label{fig:motivation}
			\vskip -0.1in
		\end{center}
		\vskip -0.2in
	\end{figure*}
	
	\textbf{Motivation 1:} 
	While augmenting LMs with topical semantics, existing approaches incorporate latent document-topic proportions and ignore an explanatory representation for each latent topic of the proportion. Here, the explanatory representation of a topic refers to a vector representation obtained from a set of high-probability terms in its topic-word distribution. For example in Figure \ref{fig:motivation}(a), we run a topic model over a document of three sentences and discover a latent document-topic proportion $\hat{\mathbf{h}}_d$ as well as three topics (top-5 key terms) correspondingly explaining each latent topic (T1, T2 and T3) of the proportion.  
	Observe that the context in sent\#2 can not resolve the meaning of the word {\it chip}.  However, introducing $\hat{\mathbf{h}}_d$ with complementary explainable topics (collections of key terms) provide an abstract (latent) and a fine granularity (explanatory) outlook, respectively.  To our knowledge, the scheme of augmenting LMs with both the latent document-topic proportion and explanatory topics remains unexplored. 
	
	\textbf{Contribution 1:} Complementing the latent document-topic proportion, we also leverage explanatory topics in augmenting LMs with topical semantics in a neural composite language modeling (NCLM) framework, consisting of a neural topic model (NTM) and a neural language model (NLM).  
	
	\textbf{Motivation 2:} A sentence in a document may have a different topical discourse than its neighboring sentences or the document itself.  Illustrated in Figure \ref{fig:motivation} (b), an NTM generates two different document-topic proportions (TP) for input document $d$ and sent\#2+sent\#3 while modeling sent\#1 in the NLM. Observe that the sent\#1 expects a topic proportion dominated by topic T3 ({\it electronics}) as in TP1; however NTM generates TP2 or TP3 due to input $d$ or sent\#2+sent\#3, respectively where both the document-topic proportions are dominated by the topic T1 about {\it marketing}. Therefore, there is need to deal with such topical discourse mismatch for each sentence in the document.       
	
	\textbf{Contribution 2:} In order to retain sentence-level topical semantics, we first extract sentence-topic association, i.e., sentence-level latent topic proportion, for each sentence using NTM and then introduce them in NLM in combination with the document-topic proportion (association). 
	
	\textbf{Contribution 3:} We evaluate the proposed NCLM framework over range of tasks such as language modeling, word sense disambiguation, document classification and information retrieval. Experimental results suggest that both the explanatory topics and sentence-topic association help in improving natural language understanding.    
	\textit{Implementation} of NCLM is available at: \url{https://github.com/YatinChaudhary/NCLM}.
	
	\begin{table*}[t]
		\vskip -0.1in
		\caption{Description of the notations used in this work} 
	\label{table:notations}
	\vskip -0.1in
	\begin{center}
		\begin{small}
			\renewcommand*{\arraystretch}{1.2}
				\resizebox{\textwidth}{!}{
					\begin{tabular}{c|l||c|l}
						\toprule
						\multicolumn{1}{c|}{\bf Notation} & \multicolumn{1}{c||}{\bf Description} & \multicolumn{1}{c|}{\bf Notation} & \multicolumn{1}{c}{\bf Description} \\ 
						\midrule
						NTM, NLM  &  Neural Topic Model, Neural Language Model  & 
						$V$, $Z$, $K$    & NLM Vocab, NTM Vocab, Number of topics   \\
						LTR, ETR  &  Latent and Explainable Topic Representations	& 
						$H$, $D_E$ & LSTM hidden size, Word embedding size   \\
						$d$, $s$, $y$  &  a document, a sentence in $d$, a word in $s$ & 
						$\mathbf{V} \in \mathbb{R}^{Z}$   & BoW representation of document $d$   \\
						$d\text{-}s$, $s\text{-}y$  & $d$ after removing $s$, $s$ after removing $y$ &
						[$\mathbf{W}, \mathbf{U} \in \mathbb{R}^{H \times Z}]$    & Decoding matrix of NTM, NLM   \\
						$\mathcal{N}$, $\boldsymbol{\epsilon} \in \mathbb{R}^{K}$  & Gaussian distribution, a sample from $\mathcal{N}$ & 
						$\mathbf{E} \in \mathbb{R}^{D_E \times Z}$    & Pre-trained word embedding matrix   \\
						$[\boldsymbol{\mu}$, $\boldsymbol{\sigma}] \in \mathbb{R}^{K}$  & mean, variance of approximate $\mathcal{N}$ & 
						[$\mathbf{r}_m$, $\mathbf{o}_m$]$ \in \mathbf{R}^{H}$ & hidden, output vector of LSTM cell  \\
						$[\mathbf{h}_{d\text{-}s}$, $\mathbf{z}_{d\text{-}s}^{att}] \in \mathbb{R}^{Z}$ & LTR, ETR representations of document $d\text{-}s$ & 
						[$\mathbf{o}_{d}^{LTA}$, $\mathbf{o}_{d}^{ETA}$, $\mathbf{o}_{d}^{LETA}$]$\in \mathbb{R}^{H}$ & Topic composition of document $d$ with $\mathbf{o}_m$\\
						$\mathbf{t} \in \mathbb{R}^{K \times topN}$  & A list of topics with topN words & 
						[$\mathbf{o}_{d,s}^{LTA}$, $\mathbf{o}_{d,s}^{ETA}$, $\mathbf{o}_{d,s}^{LETA}$]$\in \mathbb{R}^{H}$ & Topic composition of sentence $s$ with $\mathbf{o}_m$    \\
						\bottomrule
				\end{tabular}}
			\end{small}
		\end{center}
		\vskip -0.2in
	\end{table*}
	
	\section{Neural Language Model}\label{sec:nlm}
	Language modeling is the task of assigning probability distribution over a sequence of words.
	Typically, language models \cite{DBLP:conf/naacl/PetersNIGCLZ18} are applied at the sentence-level.  
	Consider a sentence $s = \{(w_m,y_m)|\;\forall\;m\text{=}1\text{:}M\}$ of length $M$ in document $d$, 
	where $(w_m,y_m)$ is a tuple containing the indices of input and output words in vocabulary of size $V$. 
	A LM computes the joint probability $p(s)$ i.e., likelihood of $s$ by a product of conditional probabilities as follows:
	\[
	p(s) = p(y_1, ..., y_{M}) = p(y_1) \prod_{m=2}^{M} p(y_m|y_{1:m-1})
	\]
	where $p(y_m|y_{1:m-1})$ is the probability of word $y_m$ conditioned on preceding context $y_{1:m-1}$.
	RNN-based LMs capture linguistic properties in their recurrent hidden state $\mathbf{r}_{m} \in \mathbb{R}^{H}$ and compute output state $\mathbf{o}_m \in \mathbb{R}^{H}$ for each $y_m$:
	\begin{equation}\label{eq:lm_context_transfer}
		\begin{aligned}
			\mathbf{o}_{m}, \mathbf{r}_{m} = f(\mathbf{r}_{m-1}, w_{m}) \ \mbox{;} \
			p(y_m|y_{1:m-1}) = p(y_m|\mathbf{o}_{m})
		\end{aligned}
	\end{equation}
	where function $f(\cdot)$ can be a standard LSTM~\citep{DBLP:journals/neco/HochreiterS97} or GRU~\citep{DBLP:conf/emnlp/ChoMGBBSB14} cell and $H$ is the number of hidden units. As illustrated in Figure \ref{fig:tm_lm_model_architecture} (c), the NLM component in our proposed NCLM framework is based on LSTM cell, i.e., $f$=$f^{LSTM}$. 
	Then, the conditional $p(y_m | {\bf o}_m)$ is computed using multinomial logistic as: 
	\begin{equation} \label{eq:lm_prediction}
		p(y_m|\mathbf{o}_m) = \frac{\exp(\mathbf{o}_m^T\mathbf{U}_{:,{y_m} + \mathbf{a}_{y_m}})}{\sum_{j=1}^{V} \exp(\mathbf{o}_m^T\mathbf{U}_{:,j} + \mathbf{a}_j)}
	\end{equation}
	where $\mathbf{U} \in \mathbb{R}^{H \times V}$, $\mathbf{a} \in \mathbb{R}^{V}$ are NLM decoding parameters and $V$ is the vocabulary size.
	Here, the input $w_m$ and output $y_m$ indices are related as $y_m$=$ w_{m+1}$. Finally, NLM computes log-likelihood $\mathcal{L}^{NLM}$  of $s$ 
	as a training objective and maximizes it:  
	\vskip -0.1in
	\begin{equation} \label{eq:lm_loss}
		\mathcal{L}^{NLM} = \log p(y_1) \sum_{m=2}^{M} \log p(y_m|\mathbf{o}_{m})
	\end{equation}
	
	\section{Neural Composite Language Model}
	While NLM captures sentence-level (short-range dependencies) linguistic properties, they tend to ignore the document-level (long-range) context across sentence boundaries. 
	\citeauthor{DBLP:conf/acl/JurafskyHQK18} \yrcite{DBLP:conf/acl/JurafskyHQK18} have shown that even by considering multiple preceding sentences as the context to predict the current word, it is often difficult to capture long-term dependencies beyond a distance of 200 words in context.  
	Therefore, a composition of NLM and NTM provides a broader document-level semantic awareness during sequence modeling leveraging document-topic proportion (association) extracted using NTM. The complementary learning leads to an improved language understanding, accounting for both sentence and document-level semantics. 
	See Table~\ref{table:notations} for description of notations used.
	
	\subsection{Neural Topic Model}
\begin{figure*}[t]
		\begin{center}
			\includegraphics[scale=1.0]{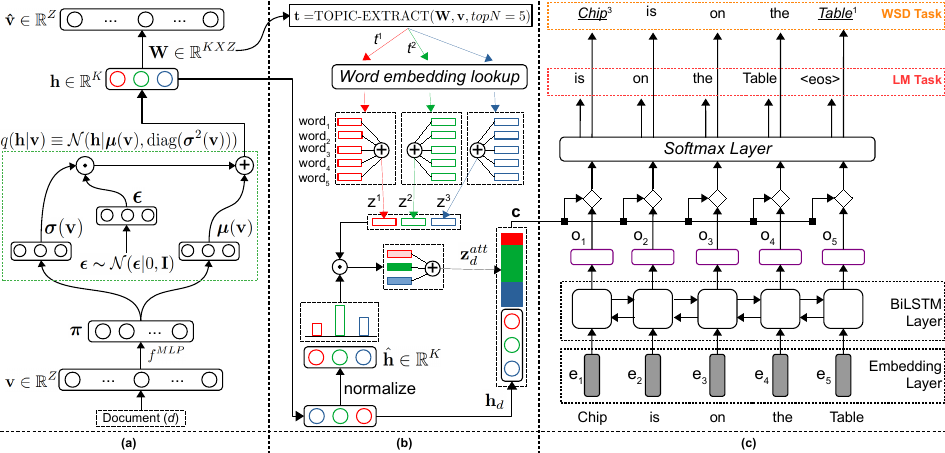}
		\end{center}
		\vskip -0.1in
		\caption{Illustration of our proposed Neural Composite Language Modeling (NCLM) framework: (a) Neural Topic Model (NTM), (b) Latent and Explainable Topic Representation Extraction and (c) Neural Language Model (NLM).}
		\label{fig:tm_lm_model_architecture}
		\vskip -0.1in
	\end{figure*}

	In this work, NTM (Figure ~\ref{fig:tm_lm_model_architecture} (a)) is based on Neural Variational Document Model (NVDM) proposed by \citeauthor{DBLP:conf/icml/MiaoYB16} \yrcite{DBLP:conf/icml/MiaoYB16}. 
	It is an unsupervised generative model that learns to regenerate 
	an input document $\mathbf{V}$ 
	using a continuous latent topic representation $\mathbf{h}$ which is sampled from a prior Gaussian distribution $p(\mathbf{h})$.
	NVDM adopts the neural variational inference framework to compute a posterior Gaussian distribution $q(\mathbf{h}|\textbf{V})$ which approximates the true prior $p(\mathbf{h})$. 
	
	Given a document d, consider $\mathbf{V} \in \mathbb{R}^{Z}$ be its bag-of-words (BoW) representation and $\mathbf{v}_i \in \mathbb{R}^{Z}$ is the one-hot representation of the $i$th word of the vocabulary of size $Z$.
	The generative process (Algorithm~\ref{algo:jointloss}: lines \#9-18) of NVDM is: 
	
	{\bf Step 1:}  Latent topic representation $\mathbf{h} \in \mathbb{R}^K$ is sampled by encoding $\mathbf{V}$ using an MLP encoder $f^{MLP}$ followed by two linear projections $l_1$ and $l_2$ as shown in Figure~\ref{fig:motivation}(a), where $\mathbf{I}$ is the identity matrix. 
	To elaborate further, for each input $\mathbf{V}$ encoder network generates the parameters mean $\boldsymbol{\mu}(\mathbf{V})$ and deviation $\boldsymbol{\sigma}(\mathbf{V})$ required to parameterize the approximate posterior distribution $q(\mathbf{h}|\mathbf{V})$ in diagonal Gaussian form and samples $\mathbf{h}$ from it (Algorithm~\ref{algo:utilityfunctions}: lines \#13-20).
	\[
	\mathbf{h} \thicksim q(\mathbf{h}|\mathbf{V}) \equiv \mathcal{N}(\mathbf{h}|\boldsymbol{\mu}(\mathbf{V}), \mbox{diag}(\boldsymbol{\sigma}^2(\mathbf{V})))
	\]
	{\bf Step 2:}  Conditional word probabilities $p(\mathbf{v}_i|\mathbf{h})$ are computed independently for each word, using multinomial logistic regression with parameters shared across all documents:
	\begin{equation}
		p(\mathbf{v}_i|\mathbf{h}) = \frac{\mbox{exp}\{\mathbf{h}^T\mathbf{W}_{:,i} + \mathbf{b}_i\}}{\sum_{j=1}^{|Z|} \mbox{exp}\{\mathbf{h}^T\mathbf{W}_{:,j} + \mathbf{b}_j\}}
	\end{equation}
	where $\mathbf{W} \in \mathbb{R}^{K \times Z}$ \& $\mathbf{b} \in \mathbb{R}^{Z}$ are NTM 
	decoding parameters.
	The word probabilities $p(\mathbf{v}_i|\mathbf{h})$ are further used to compute document probability $p(\mathbf{V}|\mathbf{h})$ conditioned on $\mathbf{h}$.
	By marginalizing $p(\mathbf{V}|\mathbf{h})$ over latent representation $\mathbf{h}$, we get the likelihood $p(\mathbf{V})$ of document $d$ as shown below.
	\[
	p(\mathbf{V}) = \int_{\mathbf{h} \thicksim p(\mathbf{h})}p(\mathbf{V}|\mathbf{h})d\mathbf{h} \quad \mbox{and} \quad p(\mathbf{V}|\mathbf{h}) = \prod_{i=1}^{N_d} p(\mathbf{v}_i|\mathbf{h})
	\]
	where $N_d$ is the number of words in document $d$.
	However, it is intractable to sample all possible configurations of $\mathbf{h} \thicksim p(\mathbf{h})$.
	Therefore, NVDM uses neural variational inference framework to compute evidence lower bound $\mathcal{L}^{NTM}$ as:
	\vskip -0.2in
	\begin{equation}\label{eq:nvdm_elbo}
		\mathcal{L}^{NTM}=\mathbb{E}_{q(\mathbf{h}|\mathbf{V})}\left[\sum_{i=1}^{N_d}\log p(\mathbf{v}_i|\mathbf{h})\right]-\mbox{KLD}
	\end{equation}
	Here $\mathcal{L}^{NTM}$ being a lower bound i.e., $\log  p(\mathbf{V}) \geq \mathcal{L}^{NTM}$, NVDM maximizes the log-likelihood of documents $\log p(\mathbf{V})$ by maximizing the evidence lower bound itself.
	The $\mathcal{L}^{NTM}$ can be maximized via back-propagation of gradients w.r.t. model parameters using the samples generated from posterior distribution $q(\mathbf{h}|\mathbf{V})$.
	NVDM assumes both prior $p(\mathbf{h})$ and posterior $q(\mathbf{h}|\mathbf{V})$ distributions as Gaussian and hence employ KL-Divergence as a regularizer term to conform $q(\mathbf{h}|\mathbf{V})$ to the Gaussian assumption i.e., $\mbox{KLD} = \mbox{KL}[q(\mathbf{h}|\mathbf{V})||p(\mathbf{h})]$, mentioned in equation~\ref{eq:nvdm_elbo}.
	
	\begin{algorithm}[t]
		\centering
		\caption{{Computation of combined loss $\mathcal{L}$}}\label{algo:jointloss}
		\begin{small}
			\begin{algorithmic}[1]
				\STATE \textbf{Input}: sentence $s = \{(w_m,y_m)|\forall m\text{=}1\text{:}M\}$ 
				\STATE \textbf{Input}: $\textbf{V} \in \mathbb{R}^{Z}$ of document $d\text{-}s$ containing $N_{d-s} words$
				\STATE \textbf{Input}: pretrained embedding matrix $\mathbf{E}$
				\STATE \textbf{Parameters}: $\{\textbf{W}, \textbf{U}, \textbf{b}, \textbf{a}, f^{MLP}, l_1, l_2, f^{LSTM}\}$
				\STATE \textbf{Hyper-parameters}: $\{\alpha, topN, g\}$
				\STATE \textbf{Initialize}: $p(\mathbf{h}) \equiv \mathcal{N}(\mathbf{h}|0,\mbox{diag}(\mathbf{I}))$
				\STATE \textbf{Initialize}: $p(\mathbf{V}|\mathbf{h}) \gets 0$; $p(s|\mathbf{V}) \gets 0$; $\mathbf{r}_0 \gets 0$
				\STATE
				\STATE \textbf{Neural Topic Model:}
				\STATE Sample Latent Topic Representation (LTR) $\mathbf{h}$
				\STATE $\mathbf{h}, \mathbf{q}(\mathbf{h}|\mathbf{V}) \gets $SAMPLE-h($f^{MLP}, g, \mathbf{V}, l_1, l_2, \mbox{sigmoid}$)
				\STATE Compute KL divergence between true prior $p(\mathbf{h})$ and $q(\mathbf{h}|\mathbf{V})$
				\STATE $\mbox{KLD} \gets \mbox{KL}[q(\mathbf{h}|\mathbf{V})||p(\mathbf{h})]$
				\FOR{$i$ from $1$ to $N_{d-s}$}
				\STATE $p(\mathbf{v}_i|\mathbf{h}) \gets \frac{\mbox{exp}\{\mathbf{h}^T\mathbf{W}_{:,i} + \mathbf{b}_i\}}{\sum_{j=1}^{Z} \mbox{exp}\{\mathbf{h}^T\mathbf{W}_{:,j} + \mathbf{b}_j\}}$
				\STATE $p(\mathbf{V}|\mathbf{h}) \gets p(\mathbf{V}|\mathbf{h}) \cdot p(\mathbf{v}_i|\mathbf{h})$
				\ENDFOR
				\STATE $\mathcal{L}^{NTM} \gets - (\log p(\mathbf{V}|\mathbf{h}) - \mbox{KLD})$
				\IF {\texttt{ETA} or \texttt{LETA}}
				\STATE Extract Explainable Topic Representation (ETR)
				\STATE $\mathbf{z}^{att}_{d\text{-}s} \gets $GET-ETR($\mathbf{W}, \mathbf{V}, topN, \mathbf{h}, \mathbf{E}$)
				\ENDIF
				\STATE 
				\STATE \textbf{Neural Composite Language Model:}
				\FOR{$m$ from $1$ to $M$}
				\STATE $\mathbf{o}_m, \mathbf{r}_m \gets f^{LSTM}(\mathbf{r}_{m-1}, w_m)$
				\STATE Composition of NTM and NLM
				\IF {\texttt{LTA}}
				\STATE $\hat{\mathbf{o}}_m \gets (\mathbf{o}_m \diamond \mathbf{h}_{d\text{-}s})$
				\ELSIF {\texttt{ETA}}
				\STATE $\hat{\mathbf{o}}_m \gets (\mathbf{o}_m \diamond \mathbf{z}^{att}_{d\text{-}s})$
				\ELSIF {\texttt{LETA}}
				\STATE $\hat{\mathbf{o}}_m \gets (\mathbf{o}_m \diamond [\mathbf{h}_{d\text{-}s}; \mathbf{z}^{att}_{d\text{-}s}]$
				\ENDIF
				\STATE $p(y_m|\mathbf{o}_m, \mathbf{V}) \gets \frac{\mbox{exp}\{\hat{\mathbf{o}}_m^T\mathbf{U}_{:,{y_m}} + \mathbf{a}_{y_m}\}}{\sum_{j=1}^{V} \mbox{exp}\{\hat{\mathbf{o}}_m^T\mathbf{U}_{:,j} + \mathbf{a}_j\}}$
				\STATE $p(s|\mathbf{V}) \gets p(s|\mathbf{V}) \cdot p(y_m|\mathbf{o}_m, \mathbf{V})$
				\ENDFOR
				\STATE $\mathcal{L}^{NLM} \gets - \log p(s|\mathbf{v})$
				\STATE $\mathcal{L} \gets \alpha \cdot \mathcal{L}^{NTM} + (1-\alpha) \cdot \mathcal{L}^{NLM}$
			\end{algorithmic}
		\end{small}
	\end{algorithm}
	
	\subsection{Topical Representation Extraction} \label{sec:topic_extraction}
	To exploit document-level semantics while language modeling, we extract topics using NVDM and represent semantics of the extracted topics in the following two forms: 
	
	\textbf{Latent Topic Extraction:} We sample latent topic representation $\mathbf{h} \in \mathbb{R}^{K}$ as shown in Figure~\ref{fig:tm_lm_model_architecture} (a) and Algorithm~\ref{algo:utilityfunctions} (lines \#13-20). Essentially, the  topic vector $\mathbf{h}$ is an 
	abstract (latent) representation of 
	topic-word distributions 
	for $K$ topics and   represents a document-topic proportion (association) as a mixture of $K$ latent topics about the document being modeled. Precisely, each scalar value $h^k \in \mathbb{R}$ denotes the \textit{contribution} of $k$th topic in representing a document $d$ by ${\bf h}$. We name $\mathbf{h}$ as {\it Latent Topic Representation} (LTR) and denote it as $\mathbf{h}_d$ for an input document $d$.
	
	\textbf{Explainable Topic Extraction:} Beyond the latent topic-proportion, we also extract explainable topics (a fine-granularity description) that can be obtained from high probability key terms of a topic-word distribution corresponding to each latent topic $k$. 
	In doing so using NTM, we use the decoding weight parameter $\mathbf{W} \in \mathbb{R}^{K \times Z}$, i.e., a topic matrix where each $k$th row $\mathbf{W}_{k,:} \in \mathbb{R}^{Z}$ denotes a distribution over vocabulary words for $k$th topic. 
	As illustrated in Figure~\ref{fig:tm_lm_model_architecture} (b), we extract key terms for each topic using the utility {\small TOPIC-EXTRACT}.  Algorithm~\ref{algo:utilityfunctions} (lines \#1-11) and Algorithm~\ref{algo:utilityfunctions} (lines \#22-30) describe the mechanism of topic learning and extracting explainable topic using {\small GET-ETR}. 
	Observe that the utility {\small TOPIC-EXTRACT} filters out the top key-terms not appearing in the document being modeled in order to highlight the contribution of those topical words shared in topic-word distribution and the document itself. Specifically, {\small TOPIC-EXTRACT} returns $K$ lists of key terms explaining each latent topic $h_k$, i.e., $\mathbf{t} = [t^k|_{k=1:K}]$ such that $t^{k}$ had $topN$ key terms for $k$th topic. We use the mask ${\bf D}$ to apply the filter as:  
	\[\mathbf{t} = \mbox{row-argmax}[\mathbf{W}\odot \mathbf{D}]_{1:topN} \]
	where ``row-argmax'' is a function which returns indices of $topN$ values from each row of input matrix, $\odot$ is an element-wise hadamard product and $\mathbf{D} \in \mathbb{R}^{K \times Z}$ is an indicator matrix where each column $\mathbf{D}_{:,i} \in \{1^{K}\;\mathbf{if}\;v_i \neq 0;  0^{K}\;\mathbf{otherwise}\}$. 
	Now for each topic $k$, we perform embedding lookup using matrix $\mathbf{E} \in \mathbb{R}^{D_E \times Z}$ (pretrained word embeddings \cite{bojanowski2017enriching}) for each word index in $t^k$  
	and then average them to compute the explanatory topic-embedding vector $\mathbf{z}^k$ as shown below:
	\vskip -0.1in
	\[
	\mathbf{z}^k = \frac{\sum_{j=1}^{topN} \mbox{emb\_lookup}(\mathbf{E},t_j^k)}{topN}
	\]
	\vskip -0.1in
	
	Finally, we perform weighted sum of topic vectors $\mathbf{z}^k$ using document-topic proportion vector $\hat{\mathbf{h}}$ as weights to compute $\mathbf{z}^{att}$. 
	We name $\mathbf{z}_{att}$ as {\it Explainable Attentive Topic Representation} (ETR) and denote it as $\mathbf{z}^{att}_d$ for a document $d$.
	\vskip -0.2in
	\[
	\mathbf{z}^{att} = \sum_{k=1}^{K}(\mathbf{z}^k \cdot \hat{h}^{k}) \quad \mbox{and} \quad  \hat{\mathbf{h}} = \mbox{softmax}(\mathbf{h}) 
	\]
	\begin{algorithm}[t]
		\centering
		\caption{{Utility functions}}\label{algo:utilityfunctions}
		\begin{small}
			\begin{algorithmic}[1]
				\FUNCTION{GET-ETR($\mathbf{W}, \mathbf{V}, topN, \mathbf{h}, \mathbf{E}$)}
				\STATE Extract $topN$ words from each topic belonging to $d$
				\STATE $\mathbf{t} \gets \mbox{TOPIC-EXTRACT}(\mathbf{W}, \mathbf{V}, topN)$
				\STATE Embedding lookup and summation to get topic embedding
				\FOR{$k$ from $1$ to $K$}
				\STATE $\mathbf{z}^k \gets \frac{\sum_{j=1}^{topN} \mbox{emb\_lookup}(\mathbf{E},t_j^k)}{topN}$
				\ENDFOR
				\STATE Weighted sum of all topic embeddings
				\STATE $\mathbf{z}^{att} \gets \sum_{k=1}^{K}(\mathbf{z}^k \cdot \hat{h}^{k})$;\;\; $\hat{\mathbf{h}} \gets \mbox{softmax}(\mathbf{h})$
				\STATE \texttt{return} $\mathbf{z}^{att}$
				\ENDFUNCTION
				\STATE
				\FUNCTION{SAMPLE-h($f, g, \mathbf{V}, l_1, l_2, \mbox{act}$)}
				\STATE Sample $\mathbf{h}$ via gaussian distribution conditioned on $\mathbf{V}$
				\STATE $\boldsymbol{\pi} \gets \mbox{act}(f(\mathbf{V}))$\;\;\;\;\;\;; $\boldsymbol{\epsilon} \thicksim \mathcal{N}(\boldsymbol{\epsilon}|0,\mbox{diag}(\mathbf{I}))$
				\STATE $\boldsymbol{\mu}(\mathbf{V}) \gets l_1(\boldsymbol{\pi})$\;\;\;\;\;\;\;; $\boldsymbol{\sigma}(\mathbf{V}) \gets l_2(\boldsymbol{\pi})$
				\STATE $q(\mathbf{h}|\mathbf{V}) \equiv \mathcal{N}(\mathbf{h}|\boldsymbol{\mu}(\mathbf{V}),\mbox{diag}(\boldsymbol{\sigma}^2(\mathbf{V})))$
				\STATE $\mathbf{h} \gets (\boldsymbol{\mu}(\mathbf{V}) + \boldsymbol{\epsilon} \odot \boldsymbol{\sigma}(\mathbf{V})) \thicksim q(\mathbf{h}|\mathbf{V})$
				\STATE \texttt{return} $g(\mathbf{h})$, $q(\mathbf{h}|\mathbf{V})$
				\ENDFUNCTION
				\STATE
				\FUNCTION{TOPIC-EXTRACT($\mathbf{W}, \mathbf{V}, topN$)}
				\STATE Create mask matrix $\mathbf{D} \in \mathbb{R}^{K \times Z}$ initialized with 0
				\FOR{$i$ from $1$ to $Z$}
				\STATE replace all 0 with 1 in column $\mathbf{D}_{:,i}$ if the count of the $i$th word of the vocabulary is non-zero in $\mathbf{V}$
				\ENDFOR
				\STATE Take hadamard product and find $topN$ max values
				\STATE $\mathbf{t} = \mbox{row-argmax}[\mathbf{W}\odot \mathbf{D}]_{1:topN}$
				\STATE \texttt{return} $\mathbf{t}$
				\ENDFUNCTION
			\end{algorithmic}
		\end{small}
	\end{algorithm}
	\subsection{Joint Topic and Language Model}
	For simplification of notation in further sections, we drop the position index in ($w_m$, $y_m$, $\mathbf{o}_m$, $\mathbf{r}_m$) from equations \{\ref{eq:lm_context_transfer},~\ref{eq:lm_prediction},~\ref{eq:lm_loss}\} and simply refer to them as ($w$, $y$, $\mathbf{o}$, $\mathbf{r}$), since our method is independent of word positions. 
	In this section we describe the \textit{composition} of topical representation $\mathbf{c} \in \{\textbf{h}_d, \textbf{z}_d^{att}\}$ with  the output vector $\mathbf{o}$ of NLM such that NLM is aware of document-level semantics while language modeling. 
	We denote composition function by ($\mathbf{o} \diamond \mathbf{c}$), where we first concatenate the two complementary representations ($\mathbf{o}$ and $\mathbf{c}$) and then perform a projection as:
	\vskip -0.1in
	\begin{equation} \label{eq:composition}
		\hat{\mathbf{o}} = (\mathbf{o} \diamond \mathbf{c}) = \mbox{sigmoid}([\mathbf{o}; \mathbf{c}]^T \mathbf{W}^{p} + \mathbf{b}^{p})
	\end{equation}
	where $\mathbf{W}^{p} \in \mathbb{R}^{\hat{H} \times H}$ and $\mathbf{b}^{p} \in \mathbb{R}^{H}$ are projection parameters, and $\hat{H} = H + K$. We then compute prediction probability of output word $y$ using equation~\ref{eq:lm_prediction} as:
	\vskip -0.1in
	\begin{equation} \label{eq:lm_prediction}
		p(y|\mathbf{o},\mathbf{c}) = \frac{\mbox{exp}\{\hat{\mathbf{o}}^T\mathbf{U}_{:,{y}} + \mathbf{a}_{y}\}}{\sum_{j=1}^{V} \mbox{exp}\{\hat{\mathbf{o}}^T\mathbf{U}_{:,j} + \mathbf{a}_j\}}
	\end{equation}
	Using this composition scheme, we employ the two representations: LTR and ETR exclusively or in combination while performing composition within NCLM framework. Following are the proposed configurations in NCLM:  
	
	\textbf{Latent Topic-aware NLM:} Existing works \cite{DBLP:conf/acl/LauBC17, DBLP:conf/iclr/Dieng0GP17, DBLP:conf/aistats/WangGWSHPSC18} in marrying topic and language models leverage latent document-topic representation $\mathbf{h}$ to incorporate document-level semantics into sequence modeling. 
	Also, modeling in such composite setting can be tricky.  To remove the chances of NLM memorizing the next word due to input to NTM, the prior works exclude the current sentence from the document before input to NTM.
	Thus for a given document $d$ and a sentence $s$ on NLM, we compute an LTR vector $\mathbf{h}_{d\text{-}s}$ by modeling $d-s$ sentences on NTM. Then, we compose it with output vector ${\bf o}$ of NLM to obtain a representation $\mathbf{o}_{d}^{LTA}$ using equation~\ref{eq:composition}, i.e., 
	$\mathbf{o}_{d}^{LTA} = (\mathbf{o} \diamond \mathbf{h}_{d\text{-}s})$. 
	We name this scheme of composition as {\it LTA-NLM}, a baseline for our contributions. 
	
	\textbf{Explainable Topic-aware NLM:} Discussed in section \ref{sec:topic_extraction}, the ETR explains each latent topic and facilitates a fine-granularity descriptive outlook by a set of key-terms. Complementary to LTR, we use the ETR vector in composition with NLM. 
	In doing so, 
	we first compose ETR representation $\mathbf{z}^{att}_{d\text{-}s}$ of $d\text{-}s$ sentences in a document $d$ with NLM output vector $\mathbf{o}$ to obtain $\mathbf{o}_{d}^{ETA}$ using equation~\ref{eq:composition}, i.e.,
	$\mathbf{o}_{d}^{ETA} = (\mathbf{o} \diamond \mathbf{z}^{att}_{d\text{-}s})$.
	This newly composite vector $\mathbf{o}_{d}^{ETA}$ encodes fine-grained explainable topical semantics to be used in sequence modeling task.
	We name this composition as {\it ETA-NLM}. To our knowledge, none of the existing approaches of joint topic and language modeling leverage explainable topics i.e., topic-word distributions 
	into NLMs. The proposed \textit{ETA-NLM} is the first one to exploit it.
	
	\textbf{Latent and Explainable Topic-aware NLM:}  We now leverage the two complementary topical representations using the latent $\mathbf{h}_{d\text{-}s}$ and explainable $\mathbf{z}^{att}_{d\text{-}s}$ vectors jointly. 
	We concatenate them together and compose it with the output vector $\mathbf{o}$ of NLM to obtain $\mathbf{o}_{d}^{LETA}$ using equation~\ref{eq:composition}, i.e., $\mathbf{o}_{d}^{LETA} = (\mathbf{o} \diamond [\mathbf{h}_{d\text{-}s}; \mathbf{z}^{att}_{d\text{-}s}])$. 
	We name this composition as {\it LETA-NLM} due to latent and explainable topic vectors. 
	
	\subsection{Sentence-level Topical Discourse} 
	Discussed in section \ref{sec:introduction} and illustrated in Figure \ref{fig:motivation}(b), there is a need for sentence-level topics in order to avoid dominant topic mismatch.  Thus, we retain sentence-level topical discourse ({\it SDT}) by incorporating sentence-level topic association (latent and/or explainable) while modeling the sentence on NLM. 
	To avoid memorization of current word being predicted $y$, we remove it from sentence $s$ i.e., $s\text{-}y$ is input to NTM to compute its topic-proportion. Given the latent and explainable representations, we first extract sentence-level LTR  $\mathbf{h}_{s\text{-}y}$ and ETR $\mathbf{z}_{s\text{-}y}^{att}$ vectors and then concatenate these with the corresponding document-level LTR and/or ETR vectors before composing them with NLM. Following are the additional  compositions for every sentence $s$ in a document $d$: 
	
	
	{\it LTA-NLM +SDT}  :  $\mathbf{o}_{d,s}^{LTA} \;\;\;= (\mathbf{o} \diamond [\mathbf{h}_{d\text{-}s}; \mathbf{h}_{s\text{-}y}])$
	
	{\it ETA-NLM +SDT}  :  $\mathbf{o}_{d,s}^{ETA} \;\;\;= (\mathbf{o} \diamond [\mathbf{z}_{d\text{-}s}^{att}; \mathbf{z}_{s\text{-}y}^{att}])$
	
	{\it LETA-NLM +SDT}: $\mathbf{o}_{d,s}^{LETA} = (\mathbf{o} \diamond [\mathbf{h}_{d\text{-}s}; \mathbf{h}_{s\text{-}y}; \mathbf{z}_{d\text{-}s}^{att}; \mathbf{z}_{s\text{-}y}^{att}])$
	
	Similarly, these composed output vectors are used to assign probability to the output word $y$ using equation~\ref{eq:lm_prediction}.
	
	To summarize, we have presented six different configurations of our proposed NCLM framework based on the composition of different latent or explainable representations as well as document-topic and sentence-topic associations:
	
	\vskip -0.1in
	\[
	p(y|\mathbf{o},\mathbf{c}) = \frac{\mbox{exp}\{\hat{\mathbf{o}}^T\mathbf{U}_{:,{y}} + \mathbf{a}_{y}\}}{\sum_{j=1}^{V} \mbox{exp}\{\hat{\mathbf{o}}^T\mathbf{U}_{:,j} + \mathbf{a}_j\}}
	\]
	\vskip -0.1in
	where, 
	$\hat{\mathbf{o}} \in \{\mathbf{o}_{d}^{LTA}, \mathbf{o}_{d}^{ETA}, \mathbf{o}_{d}^{LETA}
	\mathbf{o}_{d,s}^{LTA}, \mathbf{o}_{d,s}^{ETA}, \mathbf{o}_{d,s}^{LETA}\}$

	
	\subsection{Training Objective} \label{sec:training_objective}
	Training of the joint topic and language model is performed by maximizing the joint log-likelihood objective $\mathcal{L}$ which is a linear combination of the log-likelihood of document $d$ via NTM and sentence $s$ via NLM i.e.,
	$\mathcal{L} = \alpha \cdot \mathcal{L}^{NTM} + (1 - \alpha) \cdot \mathcal{L}^{NLM}$
	where, $\alpha \in [0,1]$ is a hyper-parameter, maintaining a balance between NTM and NLM during joint training 
	by updating model parameters at different scales.
	
	\subsection{Computational Complexity of NCLM Framework} \label{sec:computational_complexity}
	
	\textbf{NTM component complexity}: Computational complexities of extracting latent and explainable topic representations of document $d$ are described below. 
	
	\begin{enumerate}
		\item \textbf{Latent topic extraction}: complexity of computing the latent topic representation (LTR) vector $\mathbf{h}_d \in \mathbb{R}^K$, via matrix projection on the encoder side, is given as $\mathcal{O}(KZ)$, where $K$ is number of topics.
		
		
		
		\item \textbf{Explainable topic extraction}: complexity of computing explainable attentive topic representation (ETR) vector $\mathbf{z}_d^{att}$ is given as $\mathcal{O}(KZ + K(Z \log Z + D_EtopN))$, where 
		(a) $\mathcal{O}(KZ)$ is the complexity of computing mask matrix $\mathbf{D}$ and taking hadamard product with the topic matrix $\mathbf{W}$, 
		(b) $\mathcal{O}(Z \log Z)$ is the complexity of sorting $k$th row $\mathbf{W}_{k,:}$ of topic matrix $\mathbf{W}^{K \times Z}$, and 
		(c) $\mathcal{O}(D_EtopN)$ is the complexity of extracting and adding pre-trained word embeddings, via embedding matrix $\mathbf{E}$, for $topN$ key terms of $k$th topic. 
		Therefore, at asymptotic limits, computational complexity for $\mathbf{z}_d^{att}$ becomes $\mathcal{O}(K(Z \log Z + D_EtopN))$.
		
	\end{enumerate}
	
	\textbf{NLM component complexity}: NLM component of our NCLM framework is implemented as Recurrent Neural Network using LSTM cell(s). For each input word $w_m$ of sentence $s = \{(w_m,y_m)|\;\forall\;m\text{=}1\text{:}M\}$ in document $d$, the complexity of our NLM component can be sub-divided into three parts:
	
	\begin{enumerate}
		\item \textbf{Hidden state computation}: complexity of computing output state $\mathbf{o}_m$, via LSTM cell, is given as $\mathcal{O}(H(H + HH_i))$, where $H$ is the number of hidden units of LSTM cell and $H_i$ is the size of input word embedding.
		
		\item \textbf{Topic composition}: complexity of composition ($\diamond$), via concatenation and projection, of topic representation $\mathbf{c} \in \{\mathbf{h}_d, \mathbf{z}_d^{att}, [\mathbf{h}_d; \mathbf{z}_d^{att}]\}$ with output state $\mathbf{o}_m$ of NLM is given as $\mathcal{O}((H + H_{\mathbf{c}})H)$, where $H_{\mathbf{c}} \in \{K, D_E, (K+D_E)\}$ is the size of topic representation $\mathbf{c}$.
		
		\item \textbf{Word prediction}: complexity of output word prediction via multinomial logistic regression over NLM vocabulary is given as $\mathcal{O}(HV)$, where $V$ is NLM vocabulary size.
	\end{enumerate}
	
	Therefore, combined computational complexity for all $M$ words in sentence $s$ is given as 
	\vskip -0.1in
	\[
	\mathcal{O}_s = \mathcal{O}(MH(H + H_i + H_{\mathbf{c}} + V))
	\]
	
	
	
	
	
	
	\textbf{Composite model complexity}: Based on the type of topic composition, our proposed models have different computational complexities as mentioned below:

	\textit{LTA-NLM}:\;\;  $\mathcal{O}_s + \mathcal{O}(KZ)$
	
	\textit{ETA-NLM}:\;\;  $\mathcal{O}_s + \mathcal{O}(K(Z \log Z + D_EtopN))$
	
	\textit{LETA-NLM}: $\mathcal{O}_s + \mathcal{O}(K(Z \log Z + D_EtopN))$
	
	\textbf{Sentence-level topical discourse}: For each output word $y_m$ in sentence $s$, we additionally compute latent/explainable topical representation of sentence $s$ after removing $y_m$ i.e., $s\text{-}y_m$, to maintain topical discourse across sentences in document $d$.
	Therefore, $M$ additional LTR/ETR vectors are computed for all words in sentence $s$ and hence computational complexities are significantly increased by the factor of $M$ as shown below:

	\textit{LTA-NLM +SDT}:\;\;  $\mathcal{O}_s + \mathcal{O}(MKZ)$
	
	\textit{ETA-NLM +SDT}:\;\;  $\mathcal{O}_s + \mathcal{O}(MK(Z \log Z + D_EtopN))$
	
	\textit{LETA-NLM +SDT}: $\mathcal{O}_s + \mathcal{O}(MK(Z \log Z + D_EtopN))$
	
	\begin{table}[t]
		\caption{Language Modeling Perplexity scores on three datasets under two different settings of NLM i.e., S $\rightarrow$ small-NLM and L $\rightarrow$ large-NLM. Here, (+) $\rightarrow$ augment this feature to the model in previous row,     
			NLM $\rightarrow$ \textit{LSTM-LM}, ($\ast$) $\rightarrow$ scores taken from \citet{DBLP:conf/aistats/WangGWSHPSC18}.
			Dotted line separates our baseline models from the previous works ($\ast$).
			Here, \textbf{bold} values indicate best performing proposed model in comparison to \textit{LSTM-LM} baseline, and \textbf{GAIN(\%)} indicates the improvement in performance of the same.
		}
		\label{table:lm_results}
		\centering
		\begin{center}
			\begin{small}
				\begin{sc}
					\setlength{\tabcolsep}{4pt}
					\renewcommand{\arraystretch}{1.2}
					\resizebox{.47\textwidth}{!}{
								\begin{tabular}{rrcccccc}
									\toprule
									&\multicolumn{1}{c}{\multirow{2}{*}{Model}} & \multicolumn{2}{c}{APNEWS} & \multicolumn{2}{c}{IMDB} & \multicolumn{2}{c}{BNC} \\
									\cmidrule(lr){3-4} \cmidrule(lr){5-6} \cmidrule(lr){7-8}
									& & S & L & S & L & S & L \\
									\midrule
									\multirow{7}{*}{\rotatebox{90}{\small \textbf{baselines}}} 
									&\textit{LDA+LSTM*}   & 54.83  & 50.17 & 69.62 & 62.78 & 96.38 & 87.28  \\ 
									&\textit{LCLM*}       & 54.18  & 50.63 & 67.78 & 67.86 & 87.47 & 80.68  \\ 
									&\textit{TopicRNN*}   & 54.12  & 50.01 & 66.45 & 60.14 & 93.55 & 84.12  \\ 
									&\textit{TDLM*}       & 52.65  & 48.21 & 63.82 & 58.59 & 86.43 & 80.58  \\ 
									&\textit{TCNLM*}      & 52.59  & 47.74 & 62.59 & 56.12 & 86.21 & 80.12  \\ 
									\cdashline{2-8}
									&\textit{LSTM-LM} & 64.95 & 59.28 & 72.31 & 65.54 & 106.82 & 98.78 \\
									&\textit{LTA-NLM} & 55.48 & 49.61 & 68.21 & 61.49 & 98.31 & 89.36 \\ 
									\midrule
									\multirow{5}{*}{\rotatebox{90}{\small \textbf{proposed}}} 
									&\textit{+ SDT} & 48.23 & 42.85 & 63.81 & 58.90 & 90.36 & \textbf{80.30} \\
									&\textit{ETA-NLM} & 49.34 & 43.19 & 59.20 & 51.40 & 95.62 & 87.22 \\
									&\textit{+ SDT} & 48.75 & 43.50 & 57.83 & 50.51 & 95.64 & 88.37 \\
									&\textit{LETA-NLM} & 48.33 & 43.17 & 58.10 & 52.35 & 94.78 & 86.73 \\
									&\textit{+ SDT} & \textbf{42.98} & \textbf{39.41} & \textbf{56.65} & \textbf{51.05} & \textbf{88.30} & 81.12 \\
									\midrule
									&\textbf{Gain(\%)} & 33.8 & 33.5 & 21.6 & 22.1 & 17.3 & 18.7 \\
									\bottomrule
							\end{tabular}}
						\end{sc}
					\end{small}
				\end{center}
				\vskip -0.1in
			\end{table}
			
			\section{Experiments and Results}
			
			To demonstrate the positive influence of composing 
			LTR and ETR representations in neural language modeling, we perform quantitative and qualitative evaluation of our proposed models on five NLP tasks.
			
			\subsection{Evaluation: Language Modeling} \label{subsec:language_modeling}
			
			We present experimental results of language modeling using our proposed models on APNEWS, IMDB and BNC datasets~\cite{DBLP:conf/acl/LauBC17}. 
			For NLM, we tokenize sentences and documents into words, lowercase all words and remove those words which occur less than 10 times.
			For NTM, we additionally remove stopwords, word occuring less than 100 times and top 0.1\% most frequent words.
			We use standard language model perplexity as the evaluation measure for our proposed models.
			For data statistics and time complexity of experiments refer appendix.
			
			\textbf{Experimental setup:} We follow~\citet{DBLP:conf/aistats/WangGWSHPSC18} for our experimental setup.
			See appendix for detailed hyperparameter settings.
			Sentence $s$ being modeled at NLM side is removed from document $d$ at NTM side.
			We use two settings of NLM component: (1) small-NLM (1-layer, 600 hidden units), and (2) large-NLM (2-layer, 900 hidden units).
			We fix the NLM sequence length to 30 and bigger sentences are split into multiple sequences of length less than 30.
			We initialize the input word embeddings for NLM with 300-dimensional pretrained embeddings extracted from \texttt{word2vec}~\cite{DBLP:journals/corr/abs-1301-3781} 
			model trained on Google News.
			We perform an ablation study to get the best setting of hyperparameters $\alpha$ and $topN$ (see appendix). 
			
			\textbf{Baselines:} We compare our proposed models with seven baseline models: 
			(i) \textit{LDA-LSTM}: concatenating pre-trained LDA topic-proportion vector with \textit{LSTM-LM}; (ii) \textit{LCLM}~\citep{DBLP:conf/acl/WangC16}; (iii) \textit{TopicRNN}~\citep{DBLP:conf/iclr/Dieng0GP17}; (iv) \textit{TDLM}~\citep{DBLP:conf/acl/LauBC17}; (v) \textit{TCNLM}~\citep{DBLP:conf/aistats/WangGWSHPSC18}; (vi) \textit{LSTM-LM}: NLM component of our proposed models; and (vii) \textit{LTA-NLM}: our baseline model.

			\textbf{Results:} Language modeling perplexity scores are presented in Table~\ref{table:lm_results}. All topic composition models outperform \textit{LSTM-LM} baseline which demonstrate the advantage of composing document topical semantics in NLM. 
			Based on the results, here are three key observations:
			(i) \textit{ETA-NLM} always performs better than \textit{LTA-NLM}
			resulting in 11\% (49.34 vs 55.48), 13.2\% (59.20 vs 68.21) and 2.7\% (95.62 vs 98.31) improvement for APNEWS, IMDB and BNC datasets respectively under small-NLM configuration.
			This behaviour asserts that ETR vector effectively captures fine-grained document topical semantics compared to LTR vector; 
			(ii) \textit{LETA-NLM} outperforms both \textit{LTA-NLM} \& \textit{ETA-NLM} by exploiting complementary semantics of ETR and LTR vectors;
			(iii) composing sentence-level topic representations i.e., \textit{+SDT}, further boost the performance by maintaining sentence-level topical discourse and hence \textit{LETA-NLM +SDT} improve upon \textit{LTA-NLM} model by 22\% (42.98 vs 55.48), 17\% (56.65 vs 68.21) and 10\% (88.30 vs 98.31) for APNEWS, IMDB and BNC datasets respectively under small-NLM configuration.
			
			\begin{table}[t]
				\caption{Top 5 words of two randomly selected topics extracted using \textit{ETA-NLM} model for APNEWS, IMDB and BNC datasets.}
				\label{table:topic_words}
				\begin{center}
					\begin{small}
						\resizebox{0.45\textwidth}{!}{
							\begin{tabular}{cccccc}
								\toprule 
								\multicolumn{2}{c}{APNEWS} & \multicolumn{2}{c}{IMDB} & \multicolumn{2}{c}{BNC} \\
								\cmidrule(lr){1-2} \cmidrule(lr){3-4} \cmidrule(lr){5-6}
								\textbf{army} & \textbf{legal} & \textbf{comedy} & \textbf{disney} & \textbf{music} & \textbf{art} \\
								\midrule
								soldiers & jurors & jokes & daffy & album & art \\
								infantry & trial & unfunny & cindrella & guitar & paintings \\
								brigade & execution & satire & alladin & band & painting \\
								veterans & jury & sandler & looney & music & museum \\
								battalion & verdict & streep & bambi & pop & gallery \\
								\bottomrule
						\end{tabular}}
					\end{small}
				\end{center}
			\end{table}
			
			
			\begin{figure}
				\begin{center}
					\centerline{\includegraphics[scale=1.0]{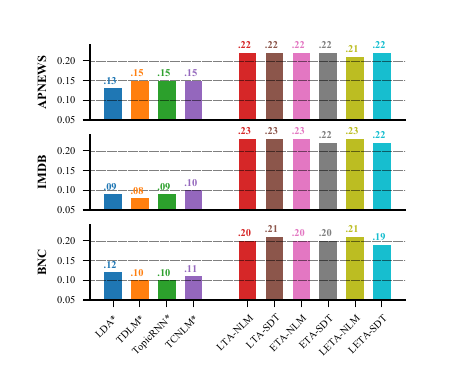}}
					\caption{Topic coherence score comparison of our proposed models and multiple baselines. ($\ast$) $\rightarrow$ taken from~\citet{DBLP:conf/aistats/WangGWSHPSC18}}
					\label{fig:TopicCoherenceBarPlot}
				\end{center}
				\vskip -0.2in
			\end{figure}
			
			\subsection{Evaluation: Topic Modeling}
			
			Typically, topic models are evaluated using perplexity measure.
			However, in the context of our NCLM framework where we compose topical semantics in language modeling, we investigate quality of topics generated by our composite models.
			We follow~\citet{DBLP:conf/aistats/WangGWSHPSC18} to infer topic coherence of top 5/10/15/20 topic words for each topic using pairwise NPMI scores and average them to get an average coherence score.
			We use the same experimental setup and hyperparameter settings as described in section~\ref{subsec:language_modeling}. 
			
			\textbf{Baselines:} Following~\citet{DBLP:conf/aistats/WangGWSHPSC18}, we compare average coherence scores of our proposed models with the following baselines: (i) LDA~\cite{DBLP:conf/nips/BleiNJ01}; (ii) TDLM~\cite{DBLP:conf/acl/LauBC17}; (iii) TopicRNN~\cite{DBLP:conf/iclr/Dieng0GP17}; (iv) TCNLM~\cite{DBLP:conf/aistats/WangGWSHPSC18}.
			
			\textbf{Results:} Average topic coherence scores are presented as bar-plot in Figure~\ref{fig:TopicCoherenceBarPlot}. 
			Based on the plot, here are two key observations:
			(i) all of our proposed models outperform every baseline by a significant margin;
			(ii) however, there is no discernible pattern in the topic coherence scores of our proposed models, hence, an improvement in language modeling performance does not correspondingly improve topic coherence. 
			For a qualitative overview, Table~\ref{table:topic_words} shows 2 randomly chosen topics  for each dataset.
			See \textit{appendix} for examples of topic-aware sentence generation.
			
			
			\begin{table}
				\caption{Text Classification accuracy scores on three datasets. \textit{CNN}$\rightarrow$model proposed by~\citet{DBLP:conf/emnlp/Kim14}, \textit{+Topic} $\rightarrow$ augment topic feature in the above model, and \textbf{bold} values indicate best models.}
				\label{table:Textclassification}
				\begin{center}
					\begin{small}
						\begin{sc}
							\renewcommand{\arraystretch}{1.2}
							\resizebox{.35\textwidth}{!}{
									\begin{tabular}{rrcccccc}
										\toprule
										& Model & 20NS & R21578 & IMDB \\ 
										\midrule
										\multirow{3}{*}{\scriptsize \rotatebox{90}{baselines}}
										&\textit{CNN-Rand}     & .721 & .690 & .888 \\
										&\textit{+Topic}       & .724 & .699 & .891 \\
										&\textit{CNN-LSTM}     & .745 & .750 & .899 \\  
										\midrule
										\multirow{3}{*}{\scriptsize \rotatebox{90}{proposed}}
										&\textit{CNN-LTA}        & .753  & .759 & .907 \\
										&\textit{CNN-ETA}        & \textbf{.775}  & \textbf{.763} & .903 \\
										&\textit{CNN-LETA}       & .770  & .750 & \textbf{.908} \\
										\bottomrule
								\end{tabular}}
							\end{sc}
						\end{small}
					\end{center}
					\vskip -0.2in
				\end{table}
				
				\subsection{Evaluation: Text Classification}
				
				We evaluate the quality of representations learned by our proposed models 
				via document classification.
				We use three labeled datasets: 20Newsgroups (20NS), Reuters (R21578) and IMDB movie reviews (IMDB) (See appendix for data statistics).
				Based on the scores in Table~\ref{table:lm_results},
				we employ our best performing composite language models as static feature extractors.
				For each document $d$, we extract: (1) output state $\mathbf{o}_m$ for each input word $x_m$ via NLM component
				; and (2) document topic representation
				vector $\mathbf{c} \in \{\mathbf{h}_d, \mathbf{z}_d^{att}, [\mathbf{h}_d; \mathbf{z}_d^{att}]\}$ via NTM component based on the model configuration.
				We then concatenate 
				$\mathbf{c}$ with each $\mathbf{o}_m$ and use them as inputs to train a CNN based text classifier proposed by~\citet{DBLP:conf/emnlp/Kim14}.
				For IMDB movie reviews dataset, we use our best models trained on unlabeled IMDB dataset as feature extractor.
				However, as 20NS and R21578 are news-domain datasets, we employ best models trained on APNEWS because of its bigger corpus size than BNC.
				
				\textbf{Baselines:} Using \textit{LTA-NLM}, \textit{ETA-NLM} and \textit{LETA-NLM} as feature extractors, we propose \textit{CNN-LTA}, \textit{CNN-ETA} and \textit{CNN-LETA} respectively.
				We compare these models with: (1) \textit{CNN-Rand}: randomly initialized CNN text classifier~\cite{DBLP:conf/emnlp/Kim14} without embedding update; (2) \textit{+Topic}: additionally concatenate LTR vector $\mathbf{h}_d$ with each word embedding input in \textit{CNN-Rand}; and (3) \textit{CNN-LSTM}: use $\mathbf{o}_m$ extracted using pre-trained \textit{LSTM-LM} as input to CNN classifier.
				
				\textbf{Results:} Document classification results are presented in Table~\ref{table:Textclassification}.
				Based on the results, there are two notable key findings: 
				(1) \textit{CNN-Rand} performed worst among all models, but incorporating document topical semantics i.e., \textit{+Topic}, provided a boost in classification scores for 20NS (.724 vs .721), R21578 (.699 vs .690) and IMDB (.891 vs .888) datasets which shows the advantage of composing document topic representations during language modeling; 
				(2) however, the best performance comes from \textit{CNN-ETA} for 20NS (.775 vs .745) \& R21578 (.763 vs .750) datasets and \textit{CNN-LETA} for IMDB (.908 vs .899).
				This shows that \textit{ETA-NLM} and \textit{LETA-NLM} learn better representations than \textit{LTA-NLM} and \textit{LSTM-LM} and suggests that ETR vector effectively captures fine-grained document semantics than LTR vector.
				
				\subsection{Evaluation: Information Retrieval}
				
				We further evaluate the quality of learned representations via document retrieval task.
				We show retrieval performance on three datasets: 20Newsgroups (20NS), Reuters (R21578) and AGnews.
				Following~\citet{DBLP:conf/aaai/GuptaCBS19}, we treat all test documents as queries and retrieve a fraction of training documents closest to each query using cosine similarity measure. 
				Then, we compute precision for each query as the fraction of all retrieved documents with same label as query and average over precision scores of all queries to get an final precision score.
				Similar to text classification, for each document $d$ of length $N_d$, we extract the final output state $\mathbf{o}_{N_d}$ of the NLM component and concatenate it with $\mathbf{c} \in \{\mathbf{h}_d, \mathbf{z}_d^{att}, [\mathbf{h}_d; \mathbf{z}_d^{att}]\}$ 
				extracted via NTM component to get a composite representation.
				We then compute cosine similarity of each query-document pair using this composite representation.
				We employ our proposed models pre-trained on APNEWS dataset as feature extractors. 
				We compute precision scores for top-5 and top-10 retrieved documents for each dataset.
				
				\textbf{Baselines:} Document retrieval is used to evaluate applicability of topic models. 
				Therefore, we compare the retrieval performance using composite representations of our best performing \textit{LTA-NLM}, \textit{ETA-NLM} and \textit{LETA-NLM} models (using table~\ref{table:lm_results}) with baseline performance using document LTR vector $\mathbf{h}_d$ extracted via pre-trained NTM component of our proposed composite language model.
				
				\textbf{Results:} Document retrieval results are presented in Table~\ref{table:InformationRetrieval}.
				It is worth noting that: 
				(1) all of our proposed composite models performed much better than the NTM component itself, and
				(2) 
				\textit{ETA-NLM} and \textit{LETA-NLM} models performed much better than \textit{LTA-NLM} which reconfirms that ETR vector is more descriptive than LTR vector and support NLM in encoding long-term semantic dependencies.
				(3) as compared to \textit{LTA-NLM}, \textit{ETA-NLM} performed best for 20NS (.376 vs .355), while \textit{LETA-NLM} performed best for R21578 (.664 vs .629) and AGnews (.694 vs .682).
				
				\begin{table}
					\caption{Information Retrieval Evaluation: Average precision scores on three labeled datasets for top-5 (P@5) \& top-10 (P@10) retrievals. \textbf{Bold} value indicates best score in each column.}
					\label{table:InformationRetrieval}
					\begin{center}
						\begin{small}
							\begin{sc}
								\renewcommand{\arraystretch}{1.1}
								\resizebox{.47\textwidth}{!}{
										\begin{tabular}{rcccccc}
											\toprule
											\multirow{2}{*}{Model} & \multicolumn{2}{c}{20NS} & \multicolumn{2}{c}{R21578} & \multicolumn{2}{c}{AGnews} \\ 
											\cmidrule(lr){2-3} \cmidrule(lr){4-5} \cmidrule(lr){6-7}
											& p@5 & p@10 & p@5 & p@10 & p@5 & p@10 \\
											\midrule
											\textit{NTM}            & .198  & .190  & .581  & .567  & .607  & .600 \\
											\textit{LTA-NLM}        & .264  & .217  & .585  & .558  & .682  & .666 \\
											\textit{ETA-NLM}        & \textbf{.287}  & \textbf{.242}  & .590  & .562  & .683  & .665 \\
											\textit{LETA-NLM}       & .281  & .236  & \textbf{.615}  & \textbf{.589}  & \textbf{.694}  & \textbf{.675} \\
											\bottomrule
									\end{tabular}}
								\end{sc}
							\end{small}
						\end{center}
						\vskip -0.2in
					\end{table}
					
					\begin{table}
						\caption{WSD evaluation results using F1 scores (micro).
							Among proposed models, \textbf{Bold} values indicates best model compared to \textit{BiLSTM-LM}.}
						\label{table:WSDscores}
						\begin{center}
							\vskip -0.1in
							\begin{small}
								\begin{sc}
									\renewcommand*{\arraystretch}{1.2}
									\resizebox{.49\textwidth}{!}{
											\begin{tabular}{rcccccc}
												\toprule
												\multicolumn{1}{c}{\multirow{2}{*}{Model}} & Dev & \multicolumn{5}{c}{Test} \\ 
												\cmidrule(lr){2-2} \cmidrule(lr){3-7}
												& SE07 & SE13  & SE15 & SE2 & SE3 & ALL \\ \hline
												\textit{MFS}     & 54.5  & 63.8 & 67.1 & 65.6  & 66.0 & 65.5 \\ 
												\textit{BiLSTM-LM}     & 55.0  & 53.9 & 60.8 & 63.6  & 60.8 & 59.8 \\ 
												\cdashline{1-7}
												\textit{LTA-NLM}        & \textbf{56.0}     &   \textbf{54.8}    &  60.9 & \textbf{64.8}  & 62.2 & \textbf{60.7}     \\
												\textit{ETA-NLM}        &    55.4    &   54.7   &     60.6  & \textbf{64.8}  & \textbf{62.3} & \textbf{60.7} \\
												\textit{LETA-NLM}       &    55.6    &   54.7   &     \textbf{61.1}  & 64.7 & 62.1 & \textbf{60.7} \\
												\bottomrule
										\end{tabular}}
									\end{sc}
								\end{small}
							\end{center}
							\vskip -0.2in
						\end{table}
						
						\subsection{Evaluation: Word Sense Disambiguation}
						Word sense disambiguation (WSD) deals with correct prediction of appropriate semantic meaning (sense) of a word given its surrounding context.
						Similar to language modeling, a word can have semantic dependencies across sentence boundaries.
						Therefore, we show the applicability of our NCLM framework which boosts 
						correct sense prediction by exploiting document-level topical knowledge to capture long-range semantic dependencies.
						We focus on English all-words WSD task where, the aim is to simultaneously predict correct sense for each word in a given sentence. 
						We use evaluation framework proposed by~\citet{DBLP:conf/eacl/NavigliCR17} for training and evaluation.
						
						\textbf{Experimental setup and Baselines:} Following~\citet{DBLP:conf/emnlp/RaganatoBN17}, we use 1-layer bidirectional-LSTM cell with 100 hidden units in the NLM component of our proposed models \textit{LTA-NLM}, \textit{ETA-NLM} and \textit{LETA-NLM}.
						In the absence of next word prediction task, 
						we use full document context on NTM side.
						Our models consider all the words in a sentence as input, and learn to predict the correct sense via multinomial logistic regression over a vocabulary of all unique senses present in training data.
						Models are trained using a learning rate of 1e-3 \& batch size of 32 and predictions are evaluated using micro F1 score.
						We compare evaluation performance of our model with the following baselines: (1) \textit{MFS}: most frequent sense extracted from WordNet~\cite{DBLP:journals/cacm/Miller95}; and (2) \textit{BiLSTM-LM}: language model using 1-layer bidirectional-LSTM cell with 100 hidden units.
						
						\textbf{Results:} 
						WSD F1 scores are presented in Table~\ref{table:WSDscores}.
						Observe that by averaging F1 scores over all test datasets, our proposed models outperform \textit{BiLSTM-LM} (60.7 vs 59.8) which again confirms the advantage of document-level semantic knowledge in resolving sense ambiguities via composition.
						
						\section{Conclusion}
						We have presented a neural composite language modeling framework that leverages both the latent and explainable topic representations 
						by composing a neural language model and a neural topic model. Moreover, we have introduced sentence-topic association along with document-topic association  
						to retain sentence-level topical discourse. Experimental results on several language understanding tasks have supported our multi-fold contributions. 
						
						\section*{Acknowledgments}
						This research was supported by Bundeswirtschaftsministerium (bmwi.de), grant 01MD19003E (PLASS (plass.io)) at Siemens AG - CT Machine Intelligence, Munich Germany.
						
						\bibliography{main}
						\bibliographystyle{icml2020}
						
						\appendix
						
						\begin{table*}[h]
							\centering
							\small
							\caption{Preprocessed dataset statistics. Here, \textbf{\#Docs} $\rightarrow$ documents, \textbf{\#Sents} $\rightarrow$ sentences, ``K'' $\rightarrow$ thousand, and ($\ast$) indicates vocabulary overlap with the corresponding vocabulary of APNEWS dataset for Information Retrieval (IR) and Text Classification tasks.}
							\label{table:datastatistics}
							\vskip 0.1in
							\renewcommand{\arraystretch}{1.3}
							\resizebox{0.85\textwidth}{!}{
								\begin{tabular}{rccccccccc}
									\toprule
									\multirow{2}{*}{\textbf{Datasets}} & \multicolumn{2}{c}{\textbf{Train}} & \multicolumn{2}{c}{\textbf{Dev}} & \multicolumn{2}{c}{\textbf{Test}} & \multicolumn{2}{c}{\textbf{Vocabulary}} & \textbf{Num} \\
									\cmidrule(lr){2-3} \cmidrule(lr){4-5} \cmidrule(lr){6-7} \cmidrule(lr){8-9}
									& \textbf{\#Docs} & \textbf{\#Sents} & \textbf{\#Docs} & \textbf{\#Sents} & \textbf{\#Docs} & \textbf{\#Sents} & \textbf{NLM} & \textbf{NTM} & \textbf{Classes} \\
									\midrule
									APNEWS & 50K & 662K & 2K & 275K & 2K & 264K & 34230 & 7990 & - \\
									IMDB & 75K & 923K & 12.5K & 153K & 12.5K & 151K & 36008 & 8714 & - \\
									BNC  & 15K & 791K & 1K & 44K & 1K & 52K & 43702 & 9741 & - \\
									\cdashline{1-10}
									20NS & 9.9K & - & 1K & - & 7.5K & - & 16884 & 6920 & 20 \\
									R21578 & 7.3K & - & 0.5K & - & 3K & - & 8650 & 4943 & 90 \\
									AGnews & 118K & - & 2K & - & 7.6K & - & 21230 & 7500 & 4  \\
									\bottomrule
							\end{tabular}}
						\end{table*}
						
						\section{Data Statistics and Evaluation}
						
						Table~\ref{table:datastatistics} shows data statistics of unlabeled and labeled datasets used to evaluate our proposed NCLM framework via Language Modeling (LM), Word Sense Disambiguation (WSD), Text Classification, Information Retrieval (IR) and Topic Modeling tasks.
						20Newsgroups (20NS), Reuters (R21578) and AGnews are news-domain datasets which are labeled with 20, 90 and 4 classes respectively. Whereas, APNEWS, BNC are unlabeled news-domain datasets. 
						However, IMDB movie reviews dataset (IMDB) is partially labeled i.e., 50K documents out of a total of 100K documents are labeled, with ``positive'' and ``negative'' sentiment labels in a single-label fashion. 
						Therefore, we utilize all 100K documents for language modeling task and 50K labeled documents for information retrieval and text classification tasks on IMDB dataset.
						For Information Retrieval (IR) and Text Classification tasks, we utilize 20NS, AGnews and R21578 datasets along with 50K labeled documents from IMDB dataset. 
						For Language Modeling (LM) and Topic Modeling (TM) tasks, we use unlabeled APNEWS, BNC, IMDB datasets and run experiments for a maximum of 100 epochs with early stopping criterion of 5 epochs.
						
						\begin{table}[htb!]
							\centering
							\caption{Hyperparameter settings of NCLM framework used in the experimental setup for Language Modeling (LM) task. Here, ($\ast$) indicates hyperparameter values taken from the experimental setup of related work as mentioned under \textbf{Experimental Setup} in subsection \textbf{4.1} in paper content.}
							\label{table:hyperparameters}
							\vskip 0.1in
							\renewcommand{\arraystretch}{1.2}
							\resizebox{.47\textwidth}{!}{
								\begin{tabular}{rc|l}
									\toprule
									& \multicolumn{1}{c}{\textbf{Hyperparameter}} & \multicolumn{1}{c}{\textbf{Value/Description}} \\
									\midrule
									\multirow{5}{*}{\rotatebox{90}{\textbf{NTM}}} & \multirow{2}{*}{$f^{MLP}$*} & \textit{1-layer} feed-forward neural network with \\
									& & 256 hidden units and \textit{sigmoid} non-linearity \\
									& $l_1, l_2$* & linear projections \\
									& $K$* & 150 \\
									& $topN$ & [10, 20, 40] \\
									& Pretraining epochs & 20 \\
									\cdashline{1-3}
									\multirow{9}{*}{\rotatebox{90}{\textbf{NLM}}} & Dropout probability* & 0.4 \\
									& Max sequence length* & 30 \\
									& small-NLM* & 1-layer LSTM-LM with 600 hidden units \\
									& large-NLM* & 2-layer LSTM-LM with 900 hidden units \\
									& Pretraining epochs* & 10 \\
									& $\alpha$ & [0.5, 0.1, 0.01] \\
									& Minibatch size* & 64 \\
									& Learning rate* & 0.001 \\
									\bottomrule
							\end{tabular}}
						\end{table}
						
						\section{Experimental setup for LM Evaluation} \label{sec:hyperparameters}
						
						Table~\ref{table:hyperparameters} shows the detailed hyperparameter settings for NTM and NLM components of our proposed NCLM framework for Language Modeling (LM) task.
						These settings are also utilized in Text Classification, Information Retrieval (IR) and Topic Modeling (TM) evaluations via best performing pretrained models on LM task.
						Based on the language modeling evaluation results from previous works for different number of topics i.e., $K \in \{50, 100, 150\}$, we fixed the number of topics $K = 150$ in the NTM component as this setting always performed best for related works.
						
						\begin{table}[htb!]
							\centering
							\caption{Ablation study over different settings of hyperparameters $\alpha$ and $topN$ for language modeling (LM) for APNEWS, IMDB and BNC datasets. For each dataset, \textbf{Bold} values indicate best LM perplexity scores and corresponding $\alpha$ \& $topN$ hyperparameter settings are finalized for extensive LM experiments.}
							\label{table:ablation_alpha}
							\vskip 0.1in
							\resizebox{.40\textwidth}{!}{
								\begin{tabular}{ccccc}
									\toprule
									& & \textbf{APNEWS} & \textbf{IMDB} & \textbf{BNC} \\ 
									\midrule
									\multirow{3}{*}{$\alpha$} & $0.5$ & 56.46 & 68.38 & 99.92 \\
									& $0.1$ & 55.97 & \textbf{68.21} & 98.85 \\
									& $0.01$ & \textbf{55.48} & 68.48 & \textbf{98.31} \\ 
									\midrule
									\multirow{3}{*}{$topN$} & $10$ & 49.26 & 60.65 & 99.63 \\
									& $20$ & \textbf{49.34} & \textbf{59.20} & 96.79 \\
									& $40$ & 51.26 & 61.95 & \textbf{95.62} \\
									\bottomrule
							\end{tabular}}
						\end{table}
						
						\section{Ablation study for $\alpha$ and $topN$} \label{sec:ablation_study}
						
						To select the best setting for hyperparameters $\alpha$ and $topN$ of our proposed NCLM framework, we perform an ablation study with $\alpha \in \{0.5, 0.1, 0.01\}$ and $topN \in \{10, 20, 40\}$ for each dataset as shown in Table~\ref{table:ablation_alpha} and select the settings with best language model perplexity scores for all of our experiments.
						We first find the best value of $\alpha$ by running experiments with LTA-NLM configuration and then freeze it to find the best value of $topN$ by running experiments with ETA-NLM configuration.
						
						\begin{table}[htb!]
							\centering
							\caption{Run-time for one epoch of our proposed models on APNEWS, IMDB, BNC datasets for language modeling task.}
							\label{table:run_times}
							\vskip 0.1in
							\resizebox{.46\textwidth}{!}{
								\begin{tabular}{rccc}
									\toprule
									\multirow{2}{*}{\textbf{Model}}& \multicolumn{3}{c}{\textbf{Run-time} (in minutes)} \\
									\cmidrule(lr){2-4}
									& \textbf{APNEWS} & \textbf{IMDB} & \textbf{BNC} \\
									\midrule
									\textit{LTA-NLM} & 45 $\pm$ 3 & 55 $\pm$ 3 & 50 $\pm$ 3 \\
									\textit{ETA-NLM} & 45 $\pm$ 3 & 55 $\pm$ 3 & 50 $\pm$ 3 \\
									\textit{LETA-NLM} & 45 $\pm$ 3 & 55 $\pm$ 3 & 50 $\pm$ 3 \\
									\textit{LTA-NLM +SDT} & 660 $\pm$ 15 & 780 $\pm$ 15 & 720 $\pm$ 15 \\
									\textit{ETA-NLM +SDT} & 660 $\pm$ 15 & 780 $\pm$ 15 & 720 $\pm$ 15 \\
									\textit{LETA-NLM +SDT} & 660 $\pm$ 15 & 780 $\pm$ 15 & 720 $\pm$ 15 \\
									\bottomrule
							\end{tabular}}
						\end{table}
						
						\section{Time complexity of NCLM configurations} 
						
						Based on the computational complexity formulation of the different configurations of our proposed NCLM framework described in section \textbf{3.6} in paper content, Table~\ref{table:run_times} shows the average run-time (in minutes) for one training epoch of our proposed models run on a NVIDIA Tesla K80 GPU with 12 GB memory.
						It is evident from Table~\ref{table:run_times} that \textit{+SDT} configurations take much more time because of the computations of LTR/ETR vectors for all $M$ words in sentence $s$.
						
						\section{Reproducibility: Code}
						
						Sections~\ref{sec:hyperparameters} and~\ref{sec:ablation_study} describe the final hyperparameter settings we used in our evaluation experiments.
						To run the experiments and reproduce the scores reported in paper content, our implementation of NCLM framework is available at \url{https://github.com/YatinChaudhary/NCLM}.
						Due to the size of model parameters and datasets beyond upload limit, we have only provided code.
						Additional information such as raw/pre-processed datasets can be obtained, as detailed in the "README.md" file.
						
						\section{Qualitative topics and Text generation}
						
						For a qualitative evaluation of topic modeling component, Table~\ref{table:topic_words} shows top 5 words of 10 selected topics extracted via NTM component for APNEWS, IMDB and BNC datasets. 
						We further investigate the text generation capability of our proposed models by generating sentences conditioned on a particular topic signal as shown in Table~\ref{table:sentencegen}.
						For a given topic $k$, during computation of latent $\mathbf{h}_d$ and explainable $\mathbf{z}_d^{att}$ topic representations of document $d$, we only utilize topic proportion/key terms of $k$th topic and suppress participation from all other topics.
						Then we use these representations and a starting token ``<bos>'' to generate sentences in a greedy token-by-token fashion.
						
						\begin{table*}
							\centering
							\caption{Top 5 words of 10 selected topics extracted from APNEWs, IDMB, BNC datasets.}
							\label{table:topic_words}
							\vskip 0.1in
							\renewcommand{\arraystretch}{1.1}
							\resizebox{0.98\textwidth}{!}{
								\begin{tabular}{rcccccccccc}
									\toprule
									\multirow{6}{*}{\rotatebox{90}{\textbf{APNEWS}}} & \textbf{ethics} & \textbf{legal} & \textbf{election} & \textbf{weather} & \textbf{music} & \textbf{fraud} & \textbf{jail} & \textbf{fire} & \textbf{festival} & \textbf{robbery} \\ 
									\cmidrule{2-11}
									& lawsuit & jurors & republicans & storm & album & fraud & inmates & flames & tourism & prison \\
									& complaint & trial & romney & winds & songs & scheme & corrections & drowned & visitors & arrested \\
									& misconduct & execution & democrats & storms & guitar & laundering & prison & engulfed & event & pleaded \\
									& violations & jury & nominee & flooding & music & fraudulent & jail & firefighters & organizers & stolen \\
									& allegations & verdict & candidates & inches & film & restitution & probation & rescuers & celebration & theft \\ 
									\midrule
									\multirow{6}{*}{\rotatebox{90}{\textbf{IMDB}}} & \textbf{fiction} & \textbf{disney} & \textbf{animation} & \textbf{comedy} & \textbf{acting} & \textbf{bollywood} & \textbf{horror} & \textbf{thriller} & \textbf{sci-fi} & \textbf{religion} \\ 
									\cmidrule{2-11}
									& spock & daffy & disney & jokes & performance & khanna & cannibal & thriller & sci-fi & muslims \\
									& batman & cindrella & animated & unfunny & supporting & saif & chainsaw & streep & alien & religion \\
									& gundam & alladin & cartoons & satire & superb & khan & leatherface & hitchcock & spaceship & muslim \\
									& superman & looney & kids & sandler & streep & amitabh & slasher & twists & science & jews \\
									& joker & bambi & anime & snl & delivers & kapoor & zombie & mystery & predator & christianity \\ 
									\midrule
									\multirow{6}{*}{\rotatebox{90}{\textbf{BNC}}} & \textbf{novel} & \textbf{health} & \textbf{pollution} & \textbf{taxation} & \textbf{art} & \textbf{family} & \textbf{sports} & \textbf{air-force} & \textbf{expression} & \textbf{business} \\ 
									\cmidrule{2-11}
									& murder & hospital & emissions & council & paintings & women & goal & aircraft & eyes & corp \\
									& police & care & environmental & cent & artist & mothers & scored & squadron & stared & ibm \\
									& book & health & recycling & million & painting & parents & players & pilot & smiled & turnover \\
									& story & nurses & waste & tax & museum & marriage & season & crew & looked & profits \\
									& detective & staff & pollution & rates & gallery & child & league & battle & shook & sales \\
									\bottomrule
							\end{tabular}}
						\end{table*}
						
						\begin{table*}
							\centering
							\caption{Examples of sentences generated via our NCLM framework under the influence of unique topic signals. Key terms explaining each topic signal are presented in Table~\ref{table:topic_words}.}
							\label{table:sentencegen}
							\vskip 0.1in
							\renewcommand*{\arraystretch}{1.3}
							\resizebox{.95\textwidth}{!}{
									\begin{tabular}{rr|l}
										\toprule
										& \multicolumn{1}{c|}{\textbf{TOPIC}} &  \multicolumn{1}{c}{\textbf{GENERATED SENTENCE}} \\ 
										\midrule
										\multirow{6}{*}{\rotatebox{90}{APNEWS}}  & \textbf{ethics} & the contract says the company will review the contract agreement with the company 's chief executive officer . \\  
										& \textbf{fraud} & prosecutors pleaded guilty in federal court in bank fraud conspiracy case . \\
										& \textbf{fire} & the fire was reported sunday night in the town of <unk> , about 20 miles northeast of los angeles . \\
										& \textbf{festival} & organizers will host events saturday at rhode island state park . \\
										& \textbf{robbery} & authorities say officers arrested 24-year-old jose <unk> in las vegas on charges of robbery and assault in \\
										&  & mexico after authorities say he shot his girlfriend in mexico in march 2012 . \\
										\cdashline{1-3}
										\multirow{7}{*}{\rotatebox{90}{IMDB}} & \textbf{thriller} & overall , it 's a solid thriller with plenty of action and action sequences , especially with a solid cast , \\
										&  & solid performances , solid action sequences . \\  
										& \textbf{comedy} & i mean , if you are trying to laugh at jokes , please avoid this crap . \\
										& \textbf{animation} & the animation is also quite impressive , but it 's not a visual achievement , but it 's a visual feast that \\
										&  & is often overlooked in its own right .  \\
										& \textbf{acting} & she plays a young woman with a strong chemistry with her character , and she plays a role with a strong performance . \\
										& \textbf{thriller} & however , it does not seem to reveal anything more than the plot , which is also quite effective . \\
										\cdashline{1-3}
										\multirow{5}{*}{\rotatebox{90}{BNC}}  & \textbf{pollution} & the <unk> is a major source of energy , and the energy supply is not a waste of energy . \\  
										& \textbf{taxation} & the chancellor 's tax cuts are not a major factor in the rise in interest rates . \\
										& \textbf{art} & the museum of art , sotheby 's , 9 june , est. \$ 150,000 -- 180,000 ; \$ 440,000 -- 180,000 ; \$ 440,000 -- <unk> ) . \\
										& \textbf{novel} & you can use the word ' <unk> ' to make a <unk> , but you can not be a detective . \\
										& \textbf{air-force} & the <unk> aircraft , which is now operational , is expected to be upgraded to <unk> <unk> . \\
										\bottomrule
								\end{tabular}}
							\end{table*}
					\end{document}